\pdfoutput=1
\documentclass[11pt]{article}
\usepackage[]{acl}
\usepackage{times}
\usepackage{latexsym}
\usepackage{multirow}
\usepackage[T1]{fontenc}

\usepackage[utf8]{inputenc}
\usepackage{booktabs}
\usepackage[normalem]{ulem}
\usepackage{microtype}
\usepackage{inconsolata}
\usepackage{graphicx,times,latexsym,float,upquote,amsmath,booktabs,microtype,footnote}
\usepackage{color,soul}
\usepackage{scrextend}

\usepackage{enumitem, tcolorbox}
\AtBeginEnvironment{tcolorbox}{\small}

\tcbset{
    sharp corners,
    colback = white,
    before skip = 0.2cm,
    after skip = 0.5cm
}                           
\newtcolorbox{boxA}{
    boxrule = 1pt,
    colframe = black 
}

\title{Automated Focused Feedback Generation for Scientific Writing Assistance}

\author{Eric Chamoun, Michael Schlichktrull , Andreas Vlachos\\
         Department of Computer Science, University of Cambridge \\
         \texttt{\{ec806,mss84,av308\}@cam.ac.uk} }

\begin{document}
\maketitle
\begin{abstract}
Scientific writing is a challenging task, particularly for novice researchers who often rely on feedback from experienced peers.  
Recent work has primarily focused on improving surface form and style rather than manuscript content. In this paper, we propose a novel task: automated focused feedback generation for scientific writing assistance\footnote{\url{https://github.com/ericchamoun/FocusedFeedbackGeneration}}. 
We present SWIF$^{2}$T: a Scientific WrIting Focused Feedback Tool. 
It is designed to generate specific, actionable and coherent comments, which identify weaknesses in a scientific paper and/or propose revisions to it.
Our approach consists of four components - planner, investigator, reviewer and controller - leveraging multiple Large Language Models (LLMs) to implement them.  
We compile a dataset of 300 peer reviews citing weaknesses in scientific papers and conduct human evaluation. The results demonstrate the superiority in specificity, reading comprehension, and overall helpfulness of SWIF$^{2}$T's feedback compared to other approaches.
In our analysis, we also identified cases where automatically generated reviews were judged better than human ones, suggesting opportunities for integration of AI-generated feedback in scientific writing.
\end{abstract}

\section{Introduction}
Peer feedback is essential in the scientific communication process, facilitating the development, interpretation, and dissemination of new findings \cite{Kuhn1962}. 
Yet, delivering high-quality feedback is both time-intensive and demanding \cite{laboriousreviews}.  
In response to this challenge, scientists are increasingly utilizing tools such as Grammarly\footnote{\url{https://www.grammarly.com}}, to improve the clarity of their manuscripts \cite{swa}.
\begin{figure}[!t]
    \centering
    \includegraphics[width = .92\linewidth]{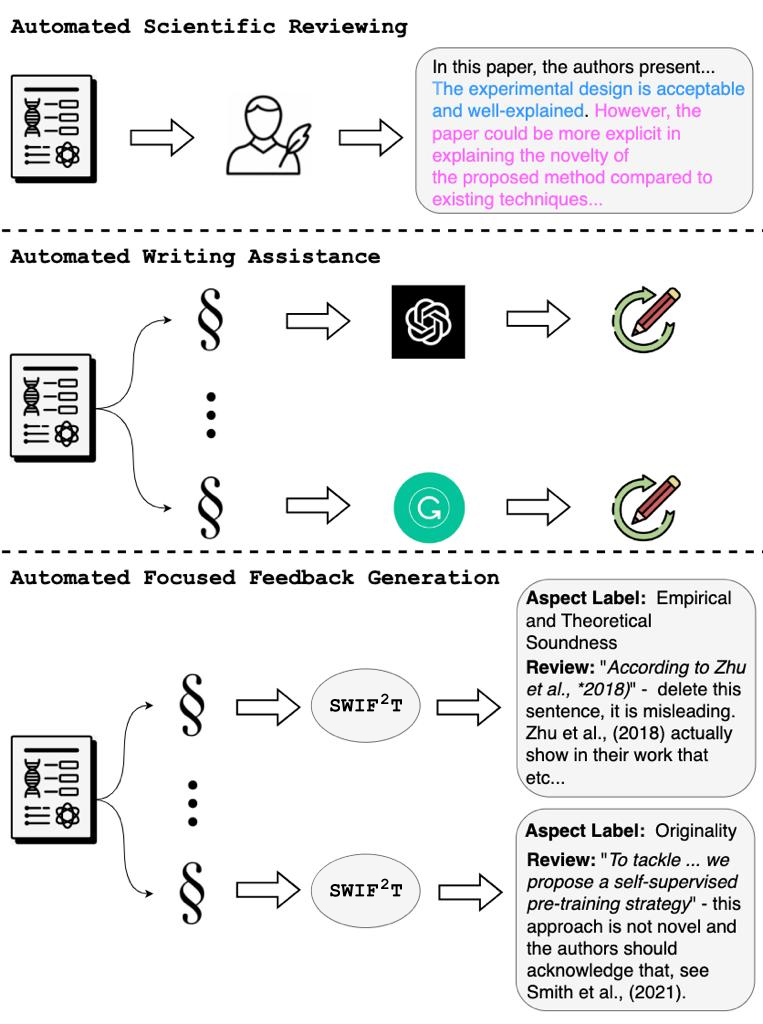}
    \caption{Overview of the proposed task. }
    \label{fig:task-overview}
\end{figure}
However, these tools focus primarily on enhancing expression clarity rather than refining the research content, and it has been argued that a scientific peer should prioritize scientific value over editorial comments to enhance the quality of a submission \cite{kelly}.

Given the importance of peer feedback in the scientific process, predicting peer response would be a powerful assistive tool for writers. Several studies \cite{reviewrobot, moprd} have explored automated scientific reviewing.
While the task is similar to scientific writing assistance, these studies focus on extracting high-level information for paper-level reviews, lacking specificity. Additionally, their emphasis is often on predicting conference acceptance rather than revising the manuscript. 
More recently, \citet{gpt4paperfeedback}
investigated the usefulness of GPT-4-generated comments on entire scientific papers. Their results indicate that, while such feedback benefits researchers, a crucial limitation lies in the models' inability to generate specific and actionable suggestions. 

In this paper, we introduce the task of automated focused feedback generation for scientific writing assistance (Figure \ref{fig:task-overview}). Focused feedback
entails providing specific, actionable, and coherent comments that identify weaknesses in a paper or suggest revisions. 
To this end, we present SWIF$^{2}$T, a Scientific WrIting Focused Feedback Tool. 
This approach leverages multiple LLMs to decompose the process of generating scientific feedback 
into four components (Figure \ref{fig:SWIFT}): planner, investigator, reviewer and controller. The planner designs a step-by-step plan involving the other components. It first instructs the investigator to answer questions using the paper or literature to enrich the paragraph under review with relevant context. Then, it directs the reviewer to leverage the gathered context to identify a specific portion requiring review, predict a weakness type (e.g.\ \textit{Originality} in Figure \ref{fig:task-overview}) before generating focused feedback accordingly. The controller manages the plan's progress and coordinates other models' actions. Recognizing the pivotal role of the plan in shaping the review outcome, we also propose a plan re-ranking method
aimed at optimizing the coherence, structure, and specificity of the generated plan, thereby enhancing the 
effectiveness of our scientific feedback generation process.

To assess our system, we compile a test set using human-written feedback derived from peer reviews. It consists of 300 examples where reviews highlight weaknesses in specific paragraphs of scientific papers. We conduct both automatic and human evaluations of SWIF$^{2}$T, comparing it against robust baselines such as CoVe \cite{cove} and GPT-4. The results demonstrate that SWIF$^{2}$T  outperforms other models not only in terms of similarity to human-written reviews, but also in specificity, reading comprehension, and overall helpfulness of its comments. Our analysis reveals instances where SWIF$^{2}$T -generated comments are preferred over human-written ones, hinting at the potential for integration of AI-generated feedback in the scientific writing process. 
Finally, our analysis highlights the strengths and weaknesses of tested LLM-based approaches, outlining avenues for future research. 

\section{Related work}

\paragraph{Scientific reviewing} In recent years, researchers have made significant contributions by constructing datasets of peer reviews for research \cite{kang,moprd,nlpeer}. Review aspects such as \textit{Originality} (Figure \ref{fig:task-overview}) have been explored in various studies, ranging from aspect score prediction \cite{kang} to aspect sentiment prediction \cite{chakraborty} and aspect-aware review generation \cite{yuan}.

These datasets have facilitated the creation of tools that automatically generate reviews from papers \citep{reviewrobot,yuan,moprd}, although these reviews tend to focus on the high-level information of a scientific paper given that the tools consume entire papers at once.  
Recently, \citet{zachary} and \citet{gpt4paperfeedback} investigated the potential of GPT-4 in assisting the peer review process. They found that GPT-4's feedback was comparable to human-written feedback in terms of helpfulness. However, they emphasized its limitations, notably its inability to provide specific, actionable feedback and to review a paper's low-level details. To address the lack of specificity in generated comments, \citet{marg} introduced MARG, a system that uses multiple LLM instances to handle different sections of scientific papers. Their proposed approach reduces generic feedback and increases comment quality. However, their focus is on generating paper-level reviews that emphasize the overall work rather than detailed aspects - which our study aims to address.

\paragraph{Text revision} Several works have developed text revision tools to 
rewrite the intended text in scientific style. A recent survey by \citet{swa} has grouped text revision tools into three different categories. The first one is Grammatical Error Correction (GEC), which seeks to detect and rectify grammatically incorrect or misspelled text 
\cite{gec}. The second category focuses on more substantial revisions than GEC, such as changing the structure of a sentence and rephrasing for clarity to generate error-free, proofread versions of early-draft sentences from scientific papers \cite{sentrev,langsmith,faltings,readreviserepeat}. The third category of text revision tools 
focuses on enhancing writing abilities through the visualization of the rhetorical structures in text \citep{movannot}. While these
approaches provide high-quality writing assistance, they do not target providing constructive criticism on the research content.

\paragraph{Scientific paper-review discourse analysis}
Recently, there has been an increased interest in understanding the interaction between scientific papers and their reviews, fueled by the availability of large open peer review datasets. 
\citet{gpt3coauthor} produced a visualization tool to study the interactions between a writer and GPT-3 when used by the former to produce real-time feedback. 

\citet{reviseresubmit} presented three tasks: pragmatic tagging, i.e. communicative purpose prediction, for classifying review sentences by whether they highlight a strength, weakness or provide a summary of the paper among other possibilities, linking to establish fine-grained relationships between review and paper sentences, and version alignment to align two revisions of the same text. 

More relevant to our research is \citet{aries}, which explores the tasks of aligning high-level feedback to specific edits and generating revisions for scientific papers.
This work differs from ours in that the presented tasks use reviews as input, whereas we focus on generating the reviews themselves. They discuss as a long-term goal the development of an ``ideal writing assistant'' that consumes relevant information from the paper and literature when generating an edit, which we implement in our work for scientific feedback generation. 

\section{Focused feedback generation} \label{sec:taskdef}
We define focused feedback as feedback that comprehends the analysed paper (\textit{reading comprehension}), clearly indicates what specific elements in the paper need revision (\textit{specificity}), and communicates to the author which actions they can take to address the issues (\textit{actionability}). These ultimately contribute to the \textit{overall helpfulness} of the feedback, 
i.e.\ the accuracy of the points discussed in the review and the extent to which they could assist the author(s) in revising their paper.

The envisioned model we aim to create concentrates on specific paragraphs $p$ within a paper $P$ to ensure no detail is missed.
Given that a paragraph on its own may lack context, the model requires access to the entire paper and relevant literature. This access 
enables the model to provide more accurate and contextualized feedback.
The desired output is comments highlighting weaknesses in the paragraph and/or suggesting revisions to the paper. 

Prior studies \cite{chakraborty,yuan,peeranalyze} utilize aspects inspired by conference reviewing guidelines\footnote{\label{refnote} Example with definitions: \url{https://acl2018.org/downloads/acl_2018_review_form.html}} to classify review sentences according to the aspects they address. In our study, we employ these aspects as weakness types to capture the focal points of reviews and exclude review sentences that do not constitute focused feedback during dataset compilation (Section \ref{sec:datacomp}). Specifically, we focus on the following five aspects\footref{refnote}: \textit{Replicability}, \textit{Originality}, \textit{Empirical and Theoretical Soundness}, \textit{Meaningful Comparison}, and \textit{Substance}. We exclude remaining aspects because they focus on refining the surface form (\textit{Readability} and \textit{Clarity}), are inherently subjective (\textit{Impact}, \textit{Motivation}) or do not constitute feedback (\textit{Summary}, \textit{Appropriateness}).

\section{SWIF$^{2}$T }

\subsection{Description}
\begin{figure}[!t]
    \centering
    \includegraphics[width = .94\linewidth]{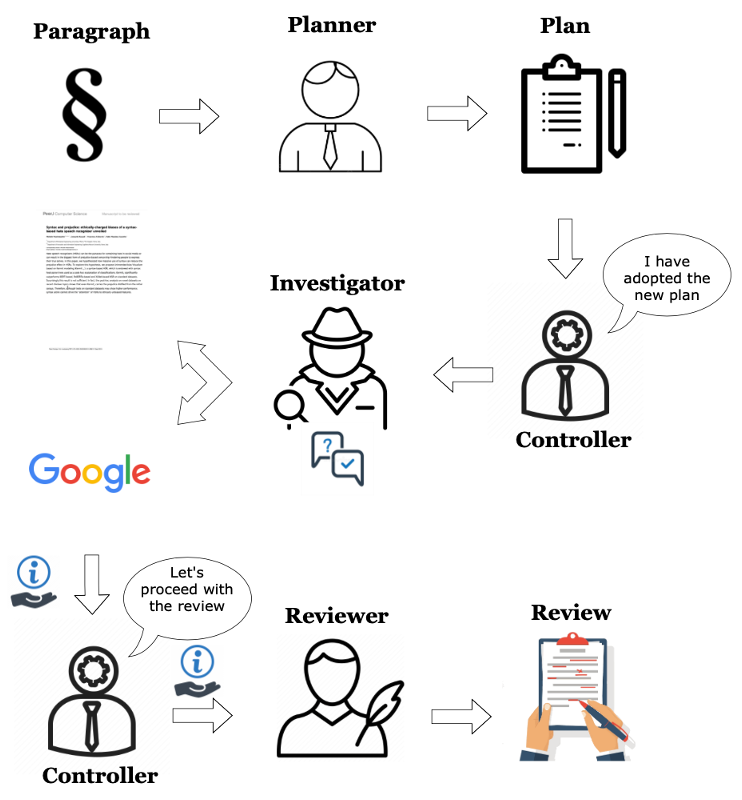}
    \caption{Illustration of our approach.
    }
    \label{fig:SWIFT}
\end{figure}

Our system (Figure \ref{fig:SWIFT}) relies on four components:  
planner, investigator, reviewer and controller. 
Given a paragraph in a paper, the planner initially formulates relevant questions and instructs the investigator to address them
using information from the rest of the paper or literature to retrieve relevant context (Figure \ref{fig:SWIFT}). 
For questions related to the literature, we use a search engine and feed retrieved pages into the investigator's QA pipeline.

\begin{figure}[!t]
    \centering
    \includegraphics[width = .9\linewidth]{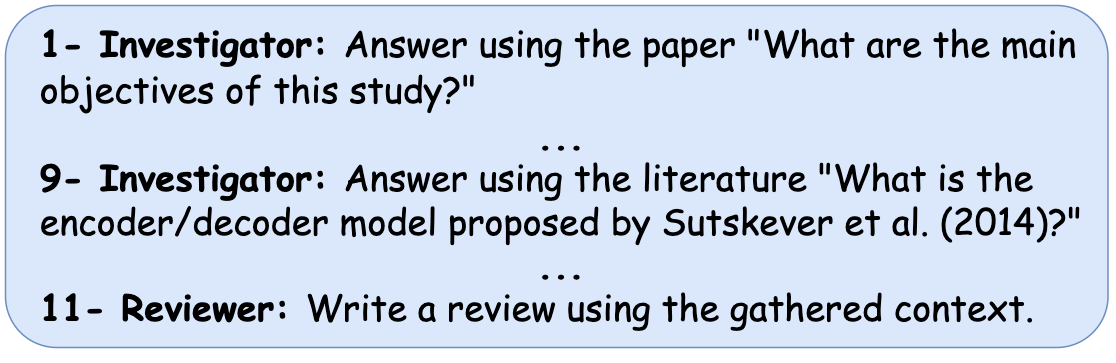}
    \caption{Example of a plan generated by SWIF$^{2}$T . We include an example of an entire run in Appendix \ref{app:run-example}.}
    \label{fig:ex-plan}
\end{figure}

To answer a question utilizing the content of the paper or a retrieved document, this pipeline first splits the document into segments of equal lengths before creating embeddings for each. A similarity search is then performed to retrieve the top 5 text chunks that are most similar to the question. Finally, our abstractive question-answering LLM generates an answer based on the retrieved paragraphs, responding with ``I don't know.'' when questions cannot be answered from the provided documents. 
Then, once enough information has been gathered, the reviewer component utilizes it to identify a section of text requiring revision, predict an aspect to concentrate the feedback on before generating the focused feedback. 
After each step in the plan (Example in Figure \ref{fig:ex-plan}), the controller decides whether the next step should be executed or replaced, enabling it to adapt the plan if needed.

\subsection{Plan re-ranking} \label{sec:planrerank}
The success of our approach relies on the quality of the generated plan according to three criteria. 
Firstly, the steps within the plan must adhere to a correct format to ensure the program's successful execution. For instance, deviations involving undefined components would lead to program failure. Secondly, the plan must exhibit coherence, ensuring that its execution results in the incorporation of relevant context from the paper and literature, thus preparing for the generation of focused feedback. Lastly, the plan steps should be specific to the input paragraph, exploring relevant concepts within it. 

We produce plans meeting this criteria by training a re-ranker to score plans w.r.t. the input paragraph.
To generate training data, we select a subset 
\textit{\textbf{S}} from our paragraph-review pairs training dataset (described in Section \ref{sec:datacomp}), where human-written reviews identify weaknesses in corresponding paragraphs.
Given a paragraph and its review, an LLM is instructed to produce a plan facilitating the generation of the review based on the input paragraph. Two instances of this LLM are configured with distinct prompts: $M_{struct}$, instructed to generate a structured plan with correctly formatted steps for successful program execution, and $M_{unstruct}$ given no structuring instruction.
For each paragraph \textbf{$p_i$} $\in$ \textbf{\textit{S}}, our objective is to generate a plan optimizing the structure, coherence, and specificity criteria. Additionally, we aim to create three alternative plans, each lacking in one of these criteria. 

Our strategy involves training a re-ranker to favor the optimal plan over alternatives, as it optimizes all criteria, including the one absent in each alternative plan.
We leverage $M_{struct}$ and $M_{unstruct}$ to generate four plans for each $p_i$:

\begin{itemize}[noitemsep, topsep=0pt, partopsep=0pt]
    \item \textbf{Optimal}: $M_{struct}$ is given as input \textbf{$p_i$} and its corresponding gold review \textbf{$r_i$} and generates a structured, coherent and specific plan to reach the review \textbf{$r_i$} from \textbf{$p_i$}.
    \item \textbf{Lacking coherence}: $M_{struct}$ is given as input \textbf{$p_i$} and the human-written review \textbf{$r_j$} of another paragraph \textbf{$p_j$} from \textit{\textbf{S}} and generates a structured, incoherent and specific plan to reach the review \textbf{$r_j$} from \textbf{$p_i$}.    
    \item \textbf{Lacking structure}: $M_{unstruct}$ is fed \textbf{$p_i$} and its corresponding human-written review \textbf{$r_i$} and generates an unstructured, coherent and specific plan to reach the review \textbf{$r_i$} from \textbf{$p_i$}.
    \item \textbf{Lacking specificity}: For each of our five weakness types, we formulate a generic comment that addresses it (e.g.\ for \textit{Originality}: ``The ideas discussed in this passage are incremental as similar ideas have already been explored in the literature.'').  These comments do not reference any concept from the paragraph and, consequently, are suitable for any paragraph that might exhibit
    such a weakness. From this set, we extract the comment with the same weakness type as $r_i$ and label it $r_{g}$. $M_{struct}$ receives \textbf{$p_i$} and  \textbf{$r_{g}$} as input, generating a structured, coherent and general plan to reach the review \textbf{$r_{g}$} from \textbf{$p_i$}.
\end{itemize}
In the training phase, the re-ranker receives a pair of inputs: one comprising a paragraph $p_i$ and the optimal plan generated for it, and the other comprising $p_i$ and one of the plans lacking in one of the desired characteristics. The re-ranker learns to score the former higher than the latter.

During inference,
the planner generates multiple plans, and the system selects the one with the highest score through pointwise inference using our trained re-ranker.

\section{Implementation}
\subsection{Dataset compilation} \label{sec:datacomp}
To construct our dataset, we utilize several existing peer review datasets that contain complete reviews and paper content: NLPeer \citep{nlpeer}, F1000rd \citep{reviseresubmit}, PeerRead \citep{kang}, MOPRD \citep{moprd}, the dataset by \citet{reviewrobot}, ASAP-Review \citep{yuan} and ARIES \cite{aries}.
We extract linked paragraph-review pairs from ARIES and a subset of F1000rd. For the remaining datasets without annotated links, we automatically retrieve all paragraph-review pairs where the review quotes a sentence within the corresponding paragraph.

We implement a two-step data filtering process to ensure our reviews (1) identify weaknesses and/or propose revisions in a paragraph while (2) providing feedback on research and technical aspects. The first step involved actionability detection: we trained a model on the subset of \textit{F1000Research} containing reviews annotated with communicative purposes (\textit{Recap}, \textit{Strength}, \textit{Todo}, \textit{Weakness}, \textit{Structure}, \textit{Other}), applied this model to our collected reviews, and only retained those indicating weaknesses (\textit{Weakness}) or suggesting recommendations (\textit{Todo}). 
Next, aspect annotation was performed on these actionable reviews. Here, we trained a model on ASAP-Review to predict the aspect addressed in our reviews from the following list: \textit{\textbf{Replicability}}, \textit{\textbf{Originality}}, \textit{\textbf{Empirical and Theoretical Soundness}}, \textit{\textbf{Meaningful Comparison}}, and \textit{\textbf{Substance}}, \textit{Summary}, \textit{Clarity} and \textit{Motivation} before then only retaining those annotated with one of the five aspects highlighted in bold as motivated in Section \ref{sec:taskdef}. By fine-tuning RoBERTa models, we achieve high accuracy in pragmatic tagging (84.40\% on test set, six labels) and aspect annotation (89.20\% on test set, eight labels), enabling effective data filtering (implementation details in Appendix \ref{app:implementationdetails}). The resulting dataset comprises 2,581 paragraphs paired with human-written reviews.

For our test set, we first randomly selected 300 examples from our dataset. Manual quality assurance was then conducted to ensure that our reviews are actionable, relevant and address one of the five aspects we focus on (Section \ref{sec:taskdef}). We replaced inadequate reviews with randomly sampled instances from the dataset that met the criteria. For instance, reviews such as \textit{Required revision in the abstract 'To our knowledge...standard manner'} (lacking specificity about the revision needed) or \textit{The statement `Types of ...' is not clear.} (misclassified as Soundness instead of Clarity)  were excluded. However, we abstained from evaluating the helpfulness of individual reviews, recognizing the subjectivity of assessing this criterion, which could introduce bias into the selected dataset.

In the final test set, 51.3\% of examples were sourced from ARIES, 20.3\% from ASAP-Review and the remainder from other discussed datasets 
 (dataset composition details in Appendix \ref{app:testdatadetails}).

\subsection{Modelling details}
We provide model and selection details in Appendix \ref{app:modeldetails}, and SWIF$^{2}$T prompts in Appendix \ref{app:prompts}.
\paragraph{Plan re-ranking}
The subset \textit{\textbf{S}} of paragraph-review pairs, used for data generation, consists of 600 examples from our training set. 
As for optimization, let $f$ be a scoring function with parameters $\theta$ that takes in a paragraph $p$ and a plan $i$ and outputs a relevance score $f_{\theta}(p,i)$ for $i$ w.r.t. $p$ according to our defined criteria. Let $J_{p^{+}}$ be the set of optimal plans w.r.t. $p \in S $ and $J_{p^{-}}$ be the set of plans lacking in one of the desired criteria w.r.t. $p$ (see Section \ref{sec:planrerank}). 
During training, the implemented re-ranker takes in a triplet consisting of paragraph $p$ and plans $i$ and $j$, and estimates the probability $P(f_{\theta}(p,i) > f_{\theta}(p,j))$ that $f$ assigns a higher score $f_{\theta}(p,i)$ to plan $i$ w.r.t $p$, compared to $f_{\theta}(p,j)$ to plan $j$ according to the defined criteria. We train the model with the following cross-entropy loss (based on \citet{duobert}):
\begin{align}
L &(\theta, (S, J)) = \sum_{p \in S; \{J_{p^{+}}, J_{p^{-}} \} \in J} \notag \\ 
-&\left[\sum_{i \in J_{p^{+}}, j \in J_{p^{-}}} \log(P(f_{\theta}(p,i) > f_{\theta}(p,j)))\right.  \\
&\Biggr.+       \log(1 - P(f_{\theta}(p,j) > f_{\theta}(p,i))) \Biggr] \notag
\end{align}
We include details about models, performance and hyperparameter search in Appendix \ref{app:implementationdetails}. In 72\% of the examples on the test set, the plan that optimizes the structure, coherence and specificity criteria is ranked the highest by our re-ranker, over plans lacking in one of these criteria. 

\paragraph{Investigator} The Google API is used to search the web. When PDFs are retrieved, Langchain's\footnote{\url{https://api.python.langchain.com/en}} PyPDF loader is used to extract the textual content. 
The LangChain API is used to split the documents into chunks of equal lengths and create an embedding for each chunk. 
To prevent our model from retrieving actual human reviews for the paragraphs -- which would not be available for a newly-written paragraph -- we  block access to peer review websites such as OpenReview, PeerJ and F1000Research.
Furthermore, given that F1000Research papers include reviews as part of the manuscripts, we manually extract relevant pages from these to avoid providing an unfair advantage to SWIF$^{2}$T in our evaluation. Our abstractive question answering model relies on providing the question-context pairs to GPT-4. It is tasked with responding with ``I don't know'' if the question cannot be answered from the given context. 

\paragraph{Reviewer} The GPT-4 based reviewer model uses in-context learning, wherein a representative training example for each weakness type is carefully chosen and incorporated into the prompt.

\section{Evaluation}
\subsection{Evaluation setup} \label{sec:evalsetup} 
Human evaluation is conducted on a subset of 100 Machine Learning and NLP examples from the test set. Each model we consider is fed a paragraph and directed to identify a portion of text containing a weakness, predict a weakness type then generate a focused review. To guide our approach to human evaluation, we conducted several pilot studies.

In shaping our evaluation criteria, insights from these pilot studies and previous research guided our selection. \citet{gpt4paperfeedback} highlighted the lack of specificity in LLM-generated reviews as a major weakness, emphasizing generic comments like ``add more experiments on more datasets.'' To address this, we compare reviews based on their \textbf{specificity and actionability}, offering an indication of how often a system provides precise suggestions that allow authors to understand the necessary actions. Secondly, pilot studies revealed a tendency for models to hallucinate in instances where context is lacking, e.g., suggesting the addition of information present in another paragraph of the paper or making unfounded assumptions. 
The \textbf{reading comprehension} criterion captures such instances. The third criterion assesses \textbf{overall helpfulness}, considering the accuracy of the point and potential impact on helping the author revise the paper. 

Additionally, the pilot studies revealed that the task's inherent complexity posed challenges for annotators in providing absolute review rankings. As a result, we opted for pairwise comparisons to mitigate subjectivity. 
No information about the approach generating a particular review was disclosed, and the order of the reviews was randomized. As there were four systems and three criteria, one example involved 18 blind pairwise comparisons - one for each criterion and each pair. In reporting our results, we opted to utilize dominance scores, which represent the sum of the instances where one approach was ranked superior to another, enhancing transparency regarding the frequency of such rankings based on specified criteria. 

Furthermore, we considered the task's difficulty and estimated the time required to annotate an example. Based on the pilot studies, we found that completing one example, i.e.,\ understanding the paragraph and reviews using the paper and literature before performing 18 pairwise comparisons, would take an average of 15 minutes. 

We recruited 11 researchers from our research group, ranging from PhD students to associate professors, all with experience in reviewing and publishing at top ML/NLP conferences. Each annotator dedicated 2.5 to 3 hours to evaluate 10 examples, performing 18 pairwise comparisons for each example. The study therefore focused on a dataset of 100 examples, with 10 being annotated twice to measure inter-annotator agreement. Dominance scores were computed based on the results of the 1,980 undertaken comparisons. The inter-annotator agreement was computed using labels for 180 comparisons from 10 examples. To ensure consistency, detailed guidelines, criteria definitions, and examples were provided to annotators through the interface (examples in Appendix \ref{app:human-eval}).
\begin{table}[H]
\centering
\resizebox{.9\linewidth}{!}{
\begin{tabular}{@{}llll@{}}
\toprule
                        & \textbf{\begin{tabular}[c]{@{}l@{}}Reading \\ comprehension\end{tabular}} & \textbf{Specificity} & \textbf{Helpfulness} \\ \midrule
\textbf{Tie \%} & 50.15                                                                     & 31.08                & 30.63                \\ \bottomrule
\end{tabular}}
\caption{Tie percentage in human evaluation test set. Reading comprehension is the criterion that annotators have found most difficult to determine.}
\end{table}

Including ties, the inter-annotator agreement is 0.3927 in Cohen $\kappa$, but rises to 0.7793 when considering only cases where one system is preferred over another. This is indicative of subjectivity in close cases, which is unsurprising given the nature of the task, but substantial agreement occurs when annotators prefer one system over another.
\subsection{Baselines}
We consider three baselines in our experiments: 

\paragraph{GPT-4 \cite{gpt4}} Similarly to \citet{gpt4paperfeedback} and \citet{zachary}, we explore the potential of GPT-4 in providing feedback on scientific papers. However, rather than operating at paper level, it takes a paragraph as input.
This model mirrors the reviewer LLM in SWIF$^{2}$T but is missing the planner, investigator and controller components. 

\paragraph{CoVe \citep{cove}} We adapt CoVe, a system designed to mitigate hallucinations. 
We modify the \textbf{Factor+Revise} version of CoVe to generate plans for collecting information as part of its broader goal of writing focused feedback. Similar to the original implementation by \citet{cove}, we employ GPT-4 for planning, answering questions, cross-checking QA pairs, and generating the review. CoVe is essentially SWIF$^{2}$T without plan re-ranking and retrieval, except it includes a step to cross-check QA pairs before feeding them into the reviewer model. This step does not constitute a major difference given that the Controller model filters out unanswered QA pairs.

\paragraph{Human-written reviews} We compare against human-written reviews. Importantly, these reviews are not treated as a ``gold standard'' but rather as a strong baseline. 
This is because we are comparing isolated snippets with model-generated feedback, which is specifically tailored for our task. The reviews were written under different conditions -- specifically, peer reviewing rather than feedback writing. Furthermore, the passages were extracted in a manner that may strip comments of their original context.

\section{Results}
\subsection{Human evaluation}
The human evaluation reveals that SWIF$^{2}$T  outperforms other baselines across all three criteria. 
\begin{table}[H]
\centering
\resizebox{.9\linewidth}{!}{
\begin{tabular}{@{}ccccc@{}}
\toprule
& \multicolumn{4}{c}{Winners}\\
 & \textbf{SWIF$^{2}$T } & \textbf{GPT-4} & \textbf{CoVe} & \textbf{Human} \\ \midrule
\textbf{SWIF$^{2}$T }                             & -                  & 22.75           & 28.90          & 29.50            \\
\textbf{GPT-4}                                 & 61.75               & -              & 48.00          & 45.75          \\
\textbf{CoVe}                                  & 50.00               & 31.50           & -             & 40.25            \\
\textbf{Human}                                  & 58.75               & 43.00             & 50.25            & -             \\ \bottomrule
\textbf{Total}                                 & \textbf{170.50}              & 97.25             & 126.25            & 115.50          \\ 
\end{tabular}}
\caption{Dominance table resulting from the pairwise comparisons in our human evaluation for the \textit{specificity and actionability} criterion. A cell with coordinates (system$_a$, system$_b$) is the sum of the number of times system $b$ was preferred over system $a$ (+1) and those when these systems were rated equally (+0.5). Scores for examples that are annotated twice are averaged. }
\end{table}
For specificity and actionability, LLM-based approaches are instructed through their prompts to extract a portion of text from the paragraph and focus the review on it. This explains the high specificity of reviews generated by LLM-based approaches, in contrast to human-written reviews that are typically shorter and more direct.
\begin{table}[H]
\centering
\resizebox{.9\linewidth}{!}{
\begin{tabular}{@{}ccccc@{}}
\toprule
& \multicolumn{4}{c}{Winners}\\
 & \textbf{SWIF$^{2}$T } & \textbf{GPT-4} & \textbf{CoVe} & \textbf{Human} \\ \midrule
\textbf{SWIF$^{2}$T }                             & -                 & 24.50           & 23.25            & 32.25          \\
\textbf{GPT-4}                                 & 52.00               & -              & 41.50          & 44.75          \\
\textbf{CoVe}                                  & 46.00               & 30.00           & -             & 41.75          \\
\textbf{Human}                                  & 45.50                 & 34.25           & 39.25          & -             \\ \bottomrule
\textbf{Total}                                 & \textbf{143.50}                 & 88.75           & 104.00            & 118.75          \\ 
\end{tabular}}
\caption{Dominance table for \textit{reading comprehension.}}
\end{table}
In reading comprehension, SWIF$^{2}$T  and human reviews perform best thanks to their access to the paper and literature. The brevity of human reviews, discussed in Section \ref{analysis}, poses challenges in gauging reading comprehension compared to LLM-generated reviews, which frequently integrate their knowledge within the review.

\begin{table}[H]
\centering
\resizebox{.9\linewidth}{!}{
\begin{tabular}{@{}ccccc@{}}
\toprule
& \multicolumn{4}{c}{Winners}\\
 & \textbf{SWIF$^{2}$T }        & \textbf{GPT-4}           & \textbf{CoVe}          & \textbf{Human}          \\ \midrule
\textbf{SWIF$^{2}$T }                             & -                         & 24.00                     & 28.00                   & 30.75                   \\

\textbf{GPT-4}                                 & 60.50                      & -                        & 52.50                   & 48.50                   \\

\textbf{CoVe}                                  & 53.25                      & 29.00                     & -                      & 42.00                   \\

\textbf{Human}                                  & 58.00                      & 39.50                   & 47.00                   & -                      \\ \midrule
\multicolumn{1}{c}{\textbf{Total} }                           & \multicolumn{1}{c}{\textbf{171.75}} & \multicolumn{1}{c}{92.50} & \multicolumn{1}{c}{127.50} & \multicolumn{1}{c}{121.25}
\end{tabular}}
\caption{Dominance table for \textit{overall helpfulness}.}
\end{table}
Finally, in terms of helpfulness, SWIF$^{2}$T  clearly outperforms other baselines while CoVe reviews are slightly preferred to human-written ones.


\subsection{Automatic evaluation}
We performed an automatic evaluation on the full test set consisting of 300 examples to assess the significance of each component in our experiments. Ablations were conducted, comparing the performance of SWIF$^{2}$T  to SWIF$^{2}$T$_{-RR}$ (SWIF$^{2}$T minus plan re-ranking) to evaluate the importance of plan re-ranking, CoVe to analyze the importance of information retrieval from the paper and literature, and vanilla GPT-4 to examine the importance of plans when compared to CoVe.

Table~\ref{tbl:aspect} shows the aspect label accuracy and F1 per category for each model w.r.t gold labels. 
\begin{table}[H]
\centering
\resizebox{\linewidth}{!}{
\begin{tabular}{@{}cccccc@{}}
\toprule
\multicolumn{1}{l}{}                                              & \multicolumn{1}{l}{}                                             & \multicolumn{1}{l}{\textbf{GPT-4}} & \multicolumn{1}{l}{\textbf{CoVe}} &  \multicolumn{1}{l}{\textbf{SWIF$^{2}$T $_{-RR}$}} &\multicolumn{1}{l}{\textbf{SWIF$^{2}$T }}  \\ \midrule
\textbf{\begin{tabular}[c]{@{}c@{}}LA\end{tabular}} &                                                                  & 24.00                               & 30.70                        & 33.00      & \textbf{34.00}                          \\
\multirow{5}{*}{\textbf{\begin{tabular}[c]{@{}c@{}}Aspect \\ F1\end{tabular}}}                               & Soundness                                                        & 30.60                               & \textbf{36.10}               & 33.51      & 31.40                                   \\
& \begin{tabular}[c]{@{}c@{}}Meaningful \\ comparison\end{tabular} & 8.70                                & 24.40                        & 31.25      & \textbf{38.00}                          \\
& Replicability                                                    & 41.30                               & 30.90                        & \textbf{45.30 }     & 42.90                          \\
& Originality                                                      & 6.60                                & 33.80                        & 35.50      & \textbf{38.90}                          \\
& Substance                                                        & 9.20                                & 24.50                        & 22.40      & \textbf{25.37}                         \\ \midrule
\end{tabular}}
\caption{Aspect label accuracy and F1 per category}
\label{tbl:aspect}
\end{table}
Aspect label accuracy results suggest that adding all the components together lead to better capturing the major weakness type of the paragraph. Notably, SWIF$^{2}$T  and SWIF$^{2}$T $_{-RR}$, which incorporate retrieval, are more effective in predicting the \textit{Meaningful comparison}, \textit{Originality} labels. This suggests that the pipeline effectively retrieves relevant context from the paper and literature, enhancing the model's awareness. Without planning, we can also see that GPT-4 performs poorly on weakness types suggesting incorporating more information or comparing to other studies or methods. In contrast, it is effective in predicting the \textit{Soundness} and \textit{Replicability} weakness types, indicating that these are less dependent on planning and retrieval. 
\begin{table}[H]
\centering
\resizebox{.45\textwidth}{!}{
\begin{tabular}{@{}cccc@{}}
\toprule
{\color[HTML]{212121} \textbf{Model}} & {\color[HTML]{212121} \textbf{METEOR}} & {\color[HTML]{212121} \textbf{BLEU@4}} & {\color[HTML]{212121} \textbf{ROUGE-L}} \\ \midrule
{\color[HTML]{212121} GPT-4}          & {\color[HTML]{212121} 18.13}           & {\color[HTML]{212121} 28.50}          & {\color[HTML]{212121} 18.37}           \\
{\color[HTML]{212121} CoVe}           & {\color[HTML]{212121} 18.76}           & {\color[HTML]{212121} 29.07}         & {\color[HTML]{212121} 19.62}           \\
{\color[HTML]{212121} SWIF$^{2}$T $_{-RR}$}      & {\color[HTML]{212121} 19.17}           & {\color[HTML]{212121} 29.77}         & {\color[HTML]{212121} 19.39 }           \\ 
{\color[HTML]{212121} SWIF$^{2}$T }      & {\color[HTML]{212121} \textbf{20.04}}           & {\color[HTML]{212121} \textbf{30.06}}         & {\color[HTML]{212121} \textbf{20.44}}           \\ \bottomrule
\end{tabular}}
\caption{Similarity of generated reviews to human reviews per model}
\label{tbl:sim}
\end{table}

Furthermore, Table~\ref{tbl:sim} shows that SWIF$^{2}$T -generated comments exhibit highest similarity to human-written ones, with a progressive increase as components are added across most metrics. 
\subsection{Analysis} \label{analysis}
We manually analyze instances of our test set used in the human evaluation to shed light on the results, utilizing examples presented in Table~\ref{tbl:examples} to explain our findings.

\paragraph{Human vs LLM-generated reviews} 
In examining examples where LLM-generated reviews were preferred to human-written ones, we aimed to examine the potential of using LLM-based approaches for scientific writing assistance. Two key observations emerged. First, instances were identified where the point made by SWIF$^{2}$T  differed but was preferred over the one in the human review (Example 1, Table ~\ref{tbl:examples}). In these cases, neither was definitively superior, and the higher-scoring review was determined based on annotator preferences, showcasing the encouraging potential of LLMs in a blind comparison setup. Secondly, when the SWIF$^{2}$T-generated and human-written reviews conveyed the same point, annotators often preferred the former, appreciating the richer content that enabled higher specificity and exhibiting contextual knowledge as opposed to the more direct nature of human-written reviews (Example 2, Table ~\ref{tbl:examples}). 
On average, comments of SWIF$^{2}$T  and CoVe consisted of 61 and 58 tokens, respectively, while GPT-4 and human-written reviews were considerably shorter at averages of 52 and 43 tokens. 

However, 
an important limitation of SWIF$^{2}$T  is its imprecision regarding studies in the literature. The literature retrieval pipeline relies on summarized information in papers to answer questions about other studies rather than on the studies themselves. In cases where human-written reviews are specific about other studies whereas SWIF$^{2}$T -generated comments are vague (``other studies'' in Example 3, Table ~\ref{tbl:examples}), the former were consistently preferred. This highlights an area for future work to further enhance SWIF$^{2}$T 's performance.

Another scenario where human reviews were favored over those generated by LLM-based approaches was in cases of hallucinations (Example 4 in Table ~\ref{tbl:examples}). While SWIF$^{2}$T  demonstrated lower susceptibility to producing generic hallucinations, it still exhibited instances of hallucinating when lacking weaknesses to discuss. This highlights the need for a weakness detection model preceding review generation to avoid such cases.

\paragraph{Comparing different LLM-based approaches} We delve into the factors influencing preferences among comments generated by our different systems. First, laying out a plan significantly enhances the likelihood of generating focused feedback.  Even CoVe, which relies solely on internal knowledge, benefits from articulating a plan as it provides background, but also a roadmap and reasoning for the review (Example 5, Table ~\ref{tbl:examples}). However, the lack of access to the paper and literature harms the performance of CoVe and GPT-4, as simple notations can confuse these models (Example 6, Table ~\ref{tbl:examples}). For this reason, SWIF$^{2}$T , which incorporates planning and has access to both the paper and literature, outperforms these models.  

\begin{table*}[h]
\resizebox{\textwidth}{!}{%
\begin{tabular}{@{}ll@{}}
\toprule
\textbf{Observation}                                                                                        & \textbf{Example}                                                           \\ \midrule
\begin{tabular}[c]{@{}l@{}}\textbf{1.} System makes a different point  than \\human-written reviews that is preferred\end{tabular}                    & \begin{tabular}[c]{@{}l@{}}\textbf{Human-written:} {\color[HTML]{656565} \begin{tabular}[c]{@{}l@{}}It's not clear which dataset was used in the experiment. \end{tabular}} \vspace{0.2cm} \\ \textbf{SWIF$^{2}$T :} {\color[HTML]{010066} \begin{tabular}[c]{@{}l@{}}The uneven distribution of individual voices, as indicated by  the statement `Individual voices are unevenly distributed, \end{tabular}}\\{\color[HTML]{010066} \begin{tabular}[c]{@{}l@{}}accounting for  from 15 minutes to over 51 hours of recorded speech', may lead to potential  biases in the  training process.\end{tabular}}\\ {\color[HTML]{010066} \begin{tabular}[c]{@{}l@{}}The authors should discuss this issue and, if possible, consider how they might address it.\end{tabular}}\end{tabular}      \\ \midrule    

{\begin{tabular}[c]{@{}l@{}}\textbf{2.} Human-written review lacking\\ specificity compared to system \\ review  making the same point\end{tabular}}                                & \begin{tabular}[c]{@{}l@{}}\textbf{Human-written:} {\color[HTML]{656565} \begin{tabular}[c]{@{}l@{}} In the introduction I'm not sure that the statement ``RL methods optimize for short-term returns'' is a fair statement.  \end{tabular}} \vspace{0.2cm} \\ \textbf{SWIF$^{2}$T :} {\color[HTML]{010066} \begin{tabular}[c]{@{}l@{}} The authors claim that 'RL methods optimize for short-term returns instead of accounting for the complete learning process, \end{tabular}}\\ {\color[HTML]{010066} \begin{tabular}[c]{@{}l@{}}  they may get stuck in local minima and fail to explore the full search space'. However, not all RL methods are short-sighted. \end{tabular}}\\ {\color[HTML]{010066} \begin{tabular}[c]{@{}l@{}} Some methods, for example, MCTS and DQN, do take the long-term reward into consideration when making decisions. \end{tabular}}\end{tabular}        \\ \midrule                                                                                                                             
\begin{tabular}[c]{@{}l@{}}\textbf{3.} Lack of preciseness of SWIF$^{2}$T \\ regarding studies from literature.\end{tabular} & \begin{tabular}[c]{@{}l@{}}\textbf{Investigator:} \textit{What criticisms or weaknesses have been pointed out in other studies regarding the use of kernel method-based theorems} \\\textit{in infinitely overparameterized regimes of neural  networks?}\\* According to https://proceedings.neurips.cc/paper/2020/file/9afe487de556e59e6db6c862adfe25a4-Paper.pdf, `` Other studies have \\ pointed out that kernel  method-based theorems require that the network weights stay very close to their initialization throughout \\ training, and cannot handle regularizers or gradient noises.'' \vspace{0.2cm} \\ \textbf{SWIF$^{2}$T :} {\color[HTML]{010066} \begin{tabular}[c]{@{}l@{}}``While the authors claim that 'kernel method-based theorems...have non-vacuous guarantees on generalization', they \end{tabular}}\\ {\color[HTML]{010066} \begin{tabular}[c]{@{}l@{}}fail to address the criticisms mentioned in other studies, particularly about the inability of such theorems to handle regularizers or\end{tabular}}\\ {\color[HTML]{010066} \begin{tabular}[c]{@{}l@{}} gradient noises.'',\end{tabular}}\end{tabular}      \\ \midrule

\textbf{4.} Confident hallucinations                                                                         & {\begin{tabular}[c]{@{}l@{}}\textbf{GPT-4:} {\color[HTML]{FD6864} \begin{tabular}[c]{@{}l@{}}The claim 'we can see in Table 8 that our proposed system correctly disambiguate its causal meaning' needs supporting evidence. \end{tabular}} \vspace{0.2cm} \\  \textbf{CoVe:} {\color[HTML]{036400} \begin{tabular}[c]{@{}l@{}}The statement 'Our experimental results show that PriorityCut outperforms state-of-the-art models in identity preservation.' lacks\end{tabular}}\\ {\color[HTML]{036400} \begin{tabular}[c]{@{}l@{}} substantial evidence or data to support this claim.\end{tabular}}\end{tabular}}                                                                                                                                                                                                                                                                                                                                                                                                                                         \\ \midrule

\textbf{5.} Usefulness of the plans                                                                                  & \begin{tabular}[c]{@{}l@{}}\underline{Relevant steps of plan}\\ \textbf{Investigator}: Answer ``\textit{Can you explain how the concept of the PDE solver works?}'' \\
\textbf{Investigator}: Answer ``\textit{What does it mean to integrate the latent space over time? How does a PDE solver achieve this?}''\\ \\ \textbf{CoVe}:{\color[HTML]{036400} \begin{tabular}[c]{@{}l@{}}  While there is a good discussion on the utilization of latent spaces in PDE solutions; such as 'With an autoencoder network architecture,\end{tabular}}\\{\color[HTML]{036400} \begin{tabular}[c]{@{}l@{}} for example, the latent space  can be used for discovering interpretable, low-dimensional dynamical models and their associated coordinates\end{tabular}}\\{\color[HTML]{036400} \begin{tabular}[c]{@{}l@{}} from high-dimensional data'. The paragraph would benefit  from a discussion or explanation of how the features or dimensions compressed\end{tabular}}\\{\color[HTML]{036400} \begin{tabular}[c]{@{}l@{}} into the latent space are selected or prioritized for different types of problems or models.\end{tabular}}\end{tabular} \\\midrule

\textbf{6.} Lack of context for CoVe and GPT-4                                                                       & \begin{tabular}[c]{@{}l@{}} \textbf{CoVe:} {\color[HTML]{036400} \begin{tabular}[c]{@{}l@{}} The sentence 'Although rigorous model selection on L is necessary to draw scientific conclusions from the results, the purpose of this \end{tabular}} \\{\color[HTML]{036400} \begin{tabular}[c]{@{}l@{}} experiment is to illustrate the utility of the CSM kernel in this application.' lacks clarity in regards to what 'L' stands for. This missing context   \end{tabular}} \\{\color[HTML]{036400} \begin{tabular}[c]{@{}l@{}} could make it hard for readers to fully comprehend the content of this sentence. \end{tabular}}\end{tabular}                                                                                                                                                                                                                                                                                                                                                                                                                                                                              \\\midrule
      
\begin{tabular}[c]{@{}l@{}} \textbf{7.} SWIF$^{2}$T  applied to our study (v1)\end{tabular}       & \begin{tabular}[c]{@{}l@{}} \textbf{Human-written: }{\color[HTML]{656565} \begin{tabular}[c]{@{}l@{}} [Section \ref{sec:datacomp}] The paper mentions compiling a dataset of 300 peer reviews citing weaknesses but provides limited details on\end{tabular}} \\{\color[HTML]{656565} \begin{tabular}[c]{@{}l@{}} the dataset's size, diversity, and sources. A more thorough explanation of the dataset's composition of the broader context. \end{tabular}} \vspace{0.2cm} \\   \textbf{SWIF$^{2}$T : }{\color[HTML]{010066} \begin{tabular}[c]{@{}l@{}} [Section \ref{sec:datacomp}] While it is clear that several existing datasets are used in construction of a new dataset, the exact specifics of how these\end{tabular}} \\{\color[HTML]{010066} \begin{tabular}[c]{@{}l@{}} comparative datasets are actually combined is not transparent enough in "... " More details on the pragmatic tagging and aspect annotation \end{tabular}} \\{\color[HTML]{010066} \begin{tabular}[c]{@{}l@{}} process would also help in ensuring replicability of the experiment.\end{tabular}} \\ \\ \textbf{Human-written: }{\color[HTML]{656565} \begin{tabular}[c]{@{}l@{}} [Section \ref{sec:evalsetup}] The paper mentioned that 11 experts were hired, with each one evaluating 10 samples. What are the criteria? \end{tabular}} \\{\color[HTML]{656565} \begin{tabular}[c]{@{}l@{}}   How to guarantee the criteria in each group are the same? How challenging is the task? \end{tabular}} \vspace{0.2cm}\\   \textbf{SWIF$^{2}$T : }{\color[HTML]{010066} \begin{tabular}[c]{@{}l@{}} [Section \ref{sec:evalsetup}] There might be room for providing more details about the annotators, such as their level of expertise. A more precise\end{tabular}} \\{\color[HTML]{010066} \begin{tabular}[c]{@{}l@{}} description of the expertise levels of the annotators is beneficial for the reader to assess the validity of the results.\end{tabular}}

\end{tabular}  

\\ \bottomrule
\end{tabular}}
\caption{Observations and examples from our manual analysis.}
\label{tbl:examples}
\end{table*}

\paragraph{Practical use case} Finally, to motivate the potential use of our task, we apply SWIF$^{2}$T  on sections \ref{sec:datacomp} and \ref{sec:evalsetup} of the previous version of our paper. The generated comments (Example 7 in Table ~\ref{tbl:examples}) align with some of the feedback received across the peer reviews, which highlights the potential of our tool in providing revisions to a paper ahead of its submission to human reviewers.

\section{Conclusion}
In this study, we introduced the task of automated focused feedback generation for scientific writing assistance. It involves generating specific, actionable and coherent comments which identify weaknesses in a paper and/or suggest revisions. We presented SWIF$^{2}$T , a system comprising planner, investigator, reviewer, and controller components. It is designed to gather context for a given paragraph and generate focused feedback. Through automatic and human evaluations, SWIF$^{2}$T  demonstrated superiority over other LLM-based approaches in similarity to human-written reviews, specificity, reading comprehension and overall helpfulness of its comments. 

\section*{Limitations}
In this study, we introduced SWIF$^{2}$T, a system for automated focused feedback in scientific writing assistance. While our model outperforms other LLM-based approaches and shows promise in aiding the feedback process, it is crucial to acknowledge limitations in real-world applications. 

\paragraph{Efficiency}
A major drawback is its cost and time intensity, relying on GPT-4 and Google queries. This limits accessibility. While the overall runtime depends on various factors including number of steps in the plan and complexity of the Google queries, SWIF$^{2}$T  runs have averaged 7 minutes and 11 seconds across the 300 examples of our dataset.

\paragraph{Usage of Google for literature search}
Using the first few pages of Google searches to answer questions about the literature has some limitations that are worth highlighting. First, using the Google API may lead to answering questions from non-scientific sources. Nonetheless, an analysis of our retrieval system's top sources indicates a predominantly scientific focus (See Appendix \ref{app:sources-retrieved}).

Additionally, this methodology might miss studies that are not well-cited or well-known, or focus on works from prestigious universities that are written in English. As a result, relevant related studies might be missed, which leads to a bias. An area of future work could involve developing a more sophisticated approach for literature-augmented question answering and context retrieval. 

\paragraph{Limitations of LLMs for feedback generation}
LLMs are biased by their training data, narrowing the scope of generated reviews. Furthermore, variability exists in the models' generations, which can lead to favorable or less favorable suggestions.

\section*{Intended uses and ethics}
Our envisioned use case for our system is to aid writers in refining drafts of scientific papers. Writers input a specific section into the system, and within minutes, receive constructive suggestions, along with structured plans and QA pairs for context and explanation. This iterative process continues until the writer is satisfied. Similar to feedback received from supervisors or peers, we expect the writer to exercise judgment in deciding whether a suggestion is worth implementing. Importantly, we want to emphasize that our tool is not intended to replace human reviewing.  Peer review serves multiple purposes beyond providing suggestions, including spotting problems and preventing manuscripts that should not be published from reaching publication stage. The aim of our tool is to assist writers in submitting the best possible paper version to a conference or journal.

\bibliography{acl_latex} 
\clearpage
\newpage
\appendix
\onecolumn
\section{Test dataset statistics} \label{app:testdatadetails}
\subsection{Details}
\begin{table}[H]
\resizebox{\textwidth}{!}{
\begin{tabular}{@{}ccc@{}}
\toprule & \textbf{Full test set}           & \textbf{Manual evaluation sample}        \\ \midrule
\textbf{Description}                                                                     & \multicolumn{2}{l}{One datapoint consists of the following entries: unique ID, paper\_id, paragraph, human\_review, human\_review\_aspect.}                                                                                                                                                        \\ \hline
\textbf{Size}                                                                            & 300                                                                                                                                                                  & 100                                                                                                                         \\ \hline
\textbf{Total number of papers}                                                          & 288                                                                                                                                                                  & 96                                                                                                                          \\ \hline
\textbf{\begin{tabular}[c]{@{}l@{}}Datapoints distribution\\ across papers\end{tabular}} & \begin{tabular}[c]{@{}l@{}}- 1 paper contributed 3 datapoints\\ - 10 papers contributed 2 datapoints each\\ - 277 papers contributed one datapoint each\end{tabular} & \begin{tabular}[c]{@{}l@{}}- 4 papers contributed 2 datapoints each\\ - 92 papers contributed 1 datapoint each\end{tabular} \\ \bottomrule
\end{tabular}}
\end{table}
\subsection{Composition}
\begin{table}[H]
\resizebox{\textwidth}{!}{
\centering
\begin{tabular}{|c|c|c|c|c|c|}
 \hline
 \multicolumn{2}{|c|}{}  & \multicolumn{2}{c}{\textbf{Full test set}}  & \multicolumn{2}{|c|}{\textbf{Sample}} 
 \\ \hline
\textbf{Dataset}                                                                                                  & \textbf{Sources}    & \textbf{Domains}                                                                                                & \textbf{\begin{tabular}[c]{@{}c@{}}Datapoints\\ from dataset\end{tabular}} & \textbf{Domains}                                                                                                & \textbf{\begin{tabular}[c]{@{}c@{}}Datapoints\\ from dataset\end{tabular}} \\ \hline
ARIES \cite{aries}                                                                & Computer Science                                                                                                  & OpenReview                                                                                                          & 154  & ML and NLP & 66                                                                               \\ \hline
ASAP-Review \cite{yuan}                                                                                                           & ML and NLP                                       & ICLR and NeurIPS                                                                                                    & 61      & ML and NLP & 25                                                                            \\ \hline
NLPeer \cite{nlpeer}                                           & \begin{tabular}[c]{@{}c@{}}ARR, COLING, CONLL,\\ F1000Research \\ excluding F1000Rd\\ annotated sample\end{tabular}                   & \begin{tabular}[c]{@{}c@{}}ML and NLP\\ Medical, Biology,\\ Metascience, Public Health\end{tabular}  & 35   & ML and NLP & 3                                                                               \\ \hline
MOPRD \cite{moprd}                                                                            & PeerJ              & \begin{tabular}[c]{@{}c@{}}Biology, Computer Science,\\ Medical, Environment,\\ Chemistry\end{tabular}                                                                                                  & 25              & ML and NLP & 3                                                                    \\ \hline
\begin{tabular}[c]{@{}c@{}}F1000Rd\\ Annotated \\ Sample \cite{reviseresubmit}\end{tabular}& F1000Research  & \begin{tabular}[c]{@{}c@{}}Medical, Biology, \\ Metascience, Public Health\end{tabular}                                                                                                                                 & 16    & N/A & 0                                                                              \\ \hline
Dataset from \citet{reviewrobot}                                                                                                                                                 & \begin{tabular}[c]{@{}c@{}}ICLR, NeurIPS,\\ ACL\end{tabular}                                                        & ML and NLP & 8 & ML and NLP & 3                                                                                  \\ \hline
PeerRead \cite{kang}                                                                                                                                                                            & ACL   & NLP                                                                                                              & 1     & N/A & 0                                                                                 \\ \hline
\end{tabular}}
\end{table}

\subsection{Extra information:} As the human evaluation was conducted by researchers from our group specializing in Machine Learning and NLP, we selected a subset of 100 datapoints within the same domain as utilized in the human evaluation. To achieve this, we initially randomly extracted 100 datapoints from our full dataset. Subsequently, we manually filtered out datapoints from domains other than Machine Learning and NLP, and randomly selected suitable replacements from the remaining dataset.
\newpage
\section{Implementation} \label{app:implementationdetails}
Model optimization was performed on the validation sets using the Huggingface hyperparameter search for communicative purpose prediction (for actionability detection) and aspect annotation \footnote{\url{https://huggingface.co/docs/transformers/hpo_ train}}. 
\begin{table}[H]
\centering
\resizebox{.5\linewidth}{!}{
\begin{tabular}{@{}cccc@{}}
\toprule
                                                                          & \begin{tabular}[c]{@{}c@{}}Communicative \\ purpose\\prediction\end{tabular} & \begin{tabular}[c]{@{}c@{}}Aspect\\ annotation\end{tabular} & \begin{tabular}[c]{@{}c@{}}Plan\\ re-ranking\end{tabular} \\ \midrule
Task                                                                      & Classification                                               & Classification                                              & Re-ranking                                                \\
Evaluation                                                                & Accuracy                                                     & Accuracy                                                    & Recall@1                                                  \\
\begin{tabular}[c]{@{}c@{}}Majority\\ baseline\\ performance\end{tabular} & 30.10\%                                                      & 12.50\%                                                     & 25.0\%                                                    \\
\begin{tabular}[c]{@{}c@{}}Model\\ performance\end{tabular}               & 84.40\%                                                      & 89.20\%                                                     & 72.0\%                                                     \\ \bottomrule
\end{tabular}}
\caption{Performance of different models on respective tasks.}
\end{table}

\begin{table}[H]
\centering
\resizebox{.6\linewidth}{!}{
\begin{tabular}{@{}cccc@{}}
\toprule
Parameter        & \begin{tabular}[c]{@{}c@{}}Communicative \\ purpose\\prediction\end{tabular} & \begin{tabular}[c]{@{}c@{}}Aspect\\ annotation\end{tabular} & \begin{tabular}[c]{@{}c@{}}Plan\\ re-ranking  \end{tabular} \\ \midrule
Checkpoint       & roberta-large                                                & roberta-large                                               & bert-base-cased                                        \\
Learning rate    & 1.8e-05                                                      & 2.3e-05                                                     & \{5e-05, 1e-06  \textbf{5e-06}, 1e-07\}                                                    \\
Batch size       & 16                                                           & 32                                                          & \{\textbf{8},16,32\}                                                         \\
Epochs & 6                                                            & 10                                                          & \textbf{7}\footnotemark                                                         \\
adam-epsilon     & 1e-08                                                        & 1e-08                                                       & 1e-08                                                     \\
Dropout          & 0.1                                                          &         0.1                                                 & \{0,\textbf{0.1}\}                                                    \\
Weight decay     & 0.01                                                         & 0.01                                                        & \{\textbf{0},0.01\}                                                         \\ \bottomrule
\end{tabular}}
\caption{Hyperparameters used during training. For plan re-ranking, tested values are reported and those leading to the best performance are in bold.}
\end{table}
\footnotetext[6]{Max 10 with early stopping}
\section{Model} \label{app:modeldetails}

All components in SWIF$^{2}$T  are based on GPT-4 (version: gpt-4-0613). 

Initially, we tested GPT-3.5 and various open-source Large Language Models (LLMs), such as Alpaca \cite{alpaca}. Similarly to \citet{zachary} who use GPT-4 in their study because they find that ``only the most powerful language models are capable of producing
coherent reviews'', we find the overall performance of other LLMs in our initial experiments unsatisfactory. Additionally, SWIF$^{2}$T  relies on multi-agent LLMs, which facilitates rapid error propagation. This affects overall effectiveness and introduces bugs. Therefore, considering that our paper introduces a new task, we utilized the most effective model available to showcase the potential of AI tools designed for focused feedback generation. 

\onecolumn
\section{Prompts} \label{app:prompts}
\subsection{Planner}
\vspace*{0.5cm}
\begin{tcolorbox}
{\fontfamily{qpl}\selectfont
\textbf{System message}
You are the ReviewGPT  Planner, a world class scientific reviewing assistant. You create plans using the Investigator and Reviewer AI agents to review paragraphs. You will ask the Investigator to gather context from both the web and the paper in the first few steps, then, at the end, the action you ask the Reviewer agent will be exactly ``Write a review based on the gathered context.''. DO NOT add a single word to this sentence. Your output MUST be formatted as a numbered list. NEVER write a step that does not involve an action for the Investigator or the Reviewer agents.}

{\fontfamily{qpl}\selectfont
\textbf{User message}
You will be given a paragraph. Your task is to point out the weaknesses of this paragraph, i.e. ask questions to gather context from the paper and the web before reasoning over it and the paragraph to identify the weaknesses of the passage. Thinking step by step, break the process of scientific reviewing down into small, simple tasks. These should involve gathering context for the paragraph, i.e., gathering information from the paper (such that the paragraph can be understood, verified and criticized without requiring any access to the paper.) and from the literature (such that the Reviewer AI agent can understand cited studies, compare the paper againt other related studies, evaluate its originality and soundness and be aware of criticisms and limitations, all without needing any access to the literature. The questions should be self-contained and formulated to facilitate effective Google searches). The gathered information should allow the Reviewer to comment on the soundness, originality, replicability, meaningfulness of the comparison or the substance of the information discussed in the paragraph. As you make plans for other AI tools, each step should be solvable using one of the following actions:

 \begin{enumerate}
     \item Actor: Investigator | Action: Answer question using the paper | Parameters: question | Description: Answer the provided question from the provided paper. It is important that the query that is searched is a question ending in `?'.
     \item Actor: Investigator | Action: Answer question using Google | Parameters: question | Description: Use google search to try to answer the provided question. It is important that the query that is searched on Google is a self-contained question ending in `?'.
     \item Actor: Reviewer | Action: Write review | Parameters: | Description: Write a review that only points out the weaknesses and areas of improvement of a passage based on the plan so far. Can only be called once context has been gathered by another agent.
 \end{enumerate}

Your plan should be a numbered list. Steps should be in simple language, and mention which agent should do them. 

I will give you now the golden rules by which you NEED to abide. It is of upmost important that none of these rules is broken: 

\noindent Rule \#1: Each step involves requesting the Investigator or the Reviewer to perform an action. 

\noindent Rule \#2: The plan begins with the Investigator answering questions using the paper. Each of these steps should start with  ``Search the paper to understand''. The next steps should request the Investigator to answer questions by searching the web. These should start with  ``Search the web to understand'' and should be only about one idea. Questions answerable from the web should be self-contained such that they are understandable without referring to another step or the paper. Finally, the last step should be EXACTLY ``Reviewer: Write a review based on the gathered context.'' 

\noindent Rule \#3: The questions to the Investigator should ONLY ask about one concept at a time. For example, ``Search the paper to understand what Attention is '' is valid but ``Search the paper to understand what Attention is and how it works'' is NOT. 

IT IS IMPORTANT TO RESPECT THE THREE GOLDEN RULES I JUST GAVE YOU. Now, the paragraph you will review is: }

\{\textit{paragraph}\}
\end{tcolorbox}

\newpage
\onecolumn
\subsection{Controller}
\vspace*{0.5cm}
\begin{tcolorbox}
{\fontfamily{qpl}\selectfont
\textbf{System message}
You are the ReviewGPT  Controller, a helpful scientific reviewing assistant. You manage several other AI agents, passing directions to them from the user. You communicate directly with the other AI agents, and as such your answers MUST be ONLY valid json.

\textbf{User message}
You are currently following an overall plan to point out the weaknesses in the paragraph: }

\{\textit{paragraph}\}

\noindent {\fontfamily{qpl}\selectfont This is a log of your progress so far: }

\{\textit{progress}\}

\noindent {\fontfamily{qpl}\selectfont The remaining steps are: }

\{\textit{steps}\}

\noindent {\fontfamily{qpl}\selectfont The next step is: }

\{\textit{next step}\}

{\fontfamily{qpl}\selectfont You will be given a list of actions. Your task is to decide what the best action to take is to accomplish the next step. Each action has several fields, separated by a vertical line (|). These are the actor who takes the action, the name of the action, the parameters that action requires, and a short description of the action. The options are:
\begin{itemize}
    \item Actor: Investigator | Action: Answer question using the paper | Parameters: question | Description: Answer the provided question from the provided paper. It is important that the query that is searched is a question ending in '?'.
    \item Actor: Investigator | Action: Answer question using Google | Parameters: question | Description: Use google search to try to answer the provided question. It is important that the query that is searched on Google is a question ending in '?'.
    \item Actor: Reviewer | Action: Write review | Parameters: | Description: Write a review that only points out the weaknesses and areas of improvement of a passage based on the plan so far. Can only be called once context has been gathered by another agent.
    \item Actor: Controller | Action: Skip this step | Parameters: | Description: Skip the current step if it is unnecessary or impossible
\end{itemize}

Provide the best action to take. Your answer must be valid JSON. It should be a JSON object with four entries, ``explanation'', ``actor'', ``action'', ``parameters''. Actors and actions should be strings, parameters should be another JSON object. Explanations should be a string containing a step-by-step description of why you chose this action. Remember, necessary parameters can be found between curly brackets in the commands. Output JUST the command.}
\end{tcolorbox}
\subsection{Investigator}
\vspace*{0.5cm}
\begin{tcolorbox}
{\fontfamily{qpl}\selectfont
\textbf{User message}
Answer the question based on the context below. IF the question cannot be answered based on the context, return exactly `I don't know'.} 

\{\textit{context}\}

{\fontfamily{qpl}\selectfont Question:} \{\textit{question}\}

{\fontfamily{qpl}\selectfont Answer:}
\end{tcolorbox}
\newpage
\subsection{Reviewer}
\vspace*{0.5cm}
\begin{tcolorbox}
{\fontfamily{qpl}\selectfont
\textbf{System message}
You are the ReviewGPT  Reviewer, a world class AI assistant for scientific reviewers. You write a review that highlights the weaknesses and areas of improvements of a paragraph based on context given to you, and return results as valid JSON. You need to make sure that the review addresses a specific portion of the paragraph, that it is not generic and that it is constructive. If you think that no review is needed, then you can also say this. Also, make sure to use the given context to generate a review: a review that points out the absence of an information in the paragraph should not be made if this information is present in another paragraph of the paper, be careful this is very important!}

{\fontfamily{qpl}\selectfont
\noindent \textbf{User message}
You will be given a paragraph with the following context:} 

\{\textit{context}\}

\noindent {\fontfamily{qpl}\selectfont There are five possible review labels: Empirical and Theoretical Soundness, Meaningful Comparison, Substance, Originality, Replicability. Write a review that: 
\begin{enumerate}
    \item Selects and quotes a substring from the given paragraph.
    \item Chooses the appropriate review label.
    \item Writes a review sentence using the quoted substring, the review label and the context. It is IMPORTANT that you use the provided context to generate a sensible review.
    \item Generates a JSON object with the keys ``reasoning'', ``label''  and ``review''. Below are examples that follow all these rules, use them as inspiration + \{in-context learning examples\} 
\end{enumerate} }
\end{tcolorbox}

\subsection{Example of SWIF$^{2}$T  run} \label{app:run-example}
\textbf{Example}

\textit{Paragraph} \footnote{Paragraphs vary in lengths from couple of sentences to full sections. However, in this case, a short paragraph lacking context was chosen to better illustrate context retrieval.}: The GF-SVM approach was deployed for the first time in the prostate cancer detection dataset, and it demonstrated a significant performance improvement over the existing models in this domain. The full features of the dataset were conducted in the experiments. 

\noindent \textit{Gold review:} The authors state (lines 205,206) that GF-SVM ``demonstrated a significant performance improvement over the existing models in this domain.'' However, the authors have failed to perform any sort of statistical analysis to demonstrate that their method is significantly better, or even different from, the established methods (e.g. see [1]).

\noindent \textbf{Plan}
\begin{enumerate}
\item  Investigator: Answer question using the paper: ``What is the GF-SVM approach?''
\item  Investigator: Answer question using the paper: ``How is the GF-SVM approach applied to the prostate cancer detection dataset?''
\item  Investigator: Answer question using the paper: ``What improvements did the GF-SVM approach bring to the existing models in prostate cancer detection?''
\item  Investigator: Answer question using the paper: ``What are the full features of the dataset used in the experiments?''
\item  Investigator: Answer question using Google: ``What are the usual performance measures in prostate cancer detection models?''
\item  Investigator: Answer question using Google: ``Are there any specific considerations or challenges in handling datasets for cancer detection?''
\item Reviewer: Write a review based on the gathered context.
\end{enumerate}

\noindent \textbf{Retrieved context}

\noindent What is the GF-SVM approach?

\noindent * The GF-SVM approach is an algorithm that combines the Genetic Folding (GF) algorithm with Support Vector Machines (SVM) for classifying patients with prostate cancer. It is a hybrid model that uses the SVM classifier with various conventional kernels to achieve high accuracy in classification.

\vspace{0.25cm}
\noindent How is the GF-SVM approach applied to the prostate cancer detection dataset?

\noindent * The GF-SVM approach is applied to the prostate cancer detection dataset by using the SVM classifier with several conventional kernels such as linear, polynomial, and RBF kernels. The performance of the GF-SVM approach is evaluated and compared to other ML approaches, and it is found to have superior accuracy.

\vspace{0.25cm}
\noindent What improvements did the GF-SVM approach bring to the existing models in prostate cancer detection?

\noindent * The GF-SVM approach brought superior accuracy performance compared to the six ML approaches in prostate cancer detection.

\vspace{0.25cm}
\noindent What are the full features of the dataset used in the experiments?

\noindent  * The full features of the dataset used in the experiments are Radius, Texture, Perimeter, Area, Smoothness, Compactness, Symmetry, Fractal\_dimension, and Diagnosis.

\vspace{0.25cm}
\noindent What are the usual performance measures in prostate cancer detection models?

\noindent * The usual performance measures in prostate cancer detection models are sensitivity, specificity, and the area under the ROC curve.

\vspace{0.25cm}
\noindent Are there any specific considerations or challenges in handling datasets for cancer detection?

\noindent * Yes, there are specific considerations and challenges in handling datasets for cancer detection, such as batch effects and the need to consider vital biological information.

\vspace{0.25cm}
\noindent \textbf{Output}

\noindent \{``review'': ``The paragraph states, `The GF-SVM approach was deployed for the first time in the prostate cancer detection dataset, and it demonstrated a significant performance improvement over the existing models in this domain.' This sentence attributes a significant performance improvement to the GF-SVM model compared to existing models but fails to provide specific performance metrics or a measure of significance. '', ``label'': ``Substance'', ``reasoning'': ``The claim of significant improvement by the GF-SVM model over existing models lacks empirical support in terms of specific metrics (sensitivity, specificity, AUC-ROC) used in this domain, as mentioned in the given context. ''\}
\section{Sources retrieved by SWIF$^{2}$T} \label{app:sources-retrieved}
\begin{table}[H]
\centering
\resizebox{.5\linewidth}{!}{
\begin{tabular}{|c|c|}
\hline
\url{aclanthology.org}           & \url{pubmed.ncbi.nlm.nih.gov}                          \\\hline
\url{nature.com}                 & \url{proceedings.mlr.press}                  \\ \hline
\url{mdpi.com}    & \url{proceedings.neurips.cc} \\ \hline
\url{arxiv.org} & \url{machinelearningmastery.com}                         \\ \hline
\url{link.springer.com}     & \url{towardsdatascience.com}                 \\ \hline
\url{en.wikipedia.org}      & \url{ncbi.nlm.nih.gov}                  \\ \hline
\multicolumn{2}{|c|}{\url{journalofbigdata.springeropen.com}}     \\ \hline                             
\end{tabular}} 
\caption{Sources retrieved at least 10 times by SWIF$^{2}$T}
\end{table}
\noindent
\onecolumn
\section{Human evaluation interface example} \label{app:human-eval}

 \begin{figure}[H]
    \centering
    \includegraphics[width = \linewidth]{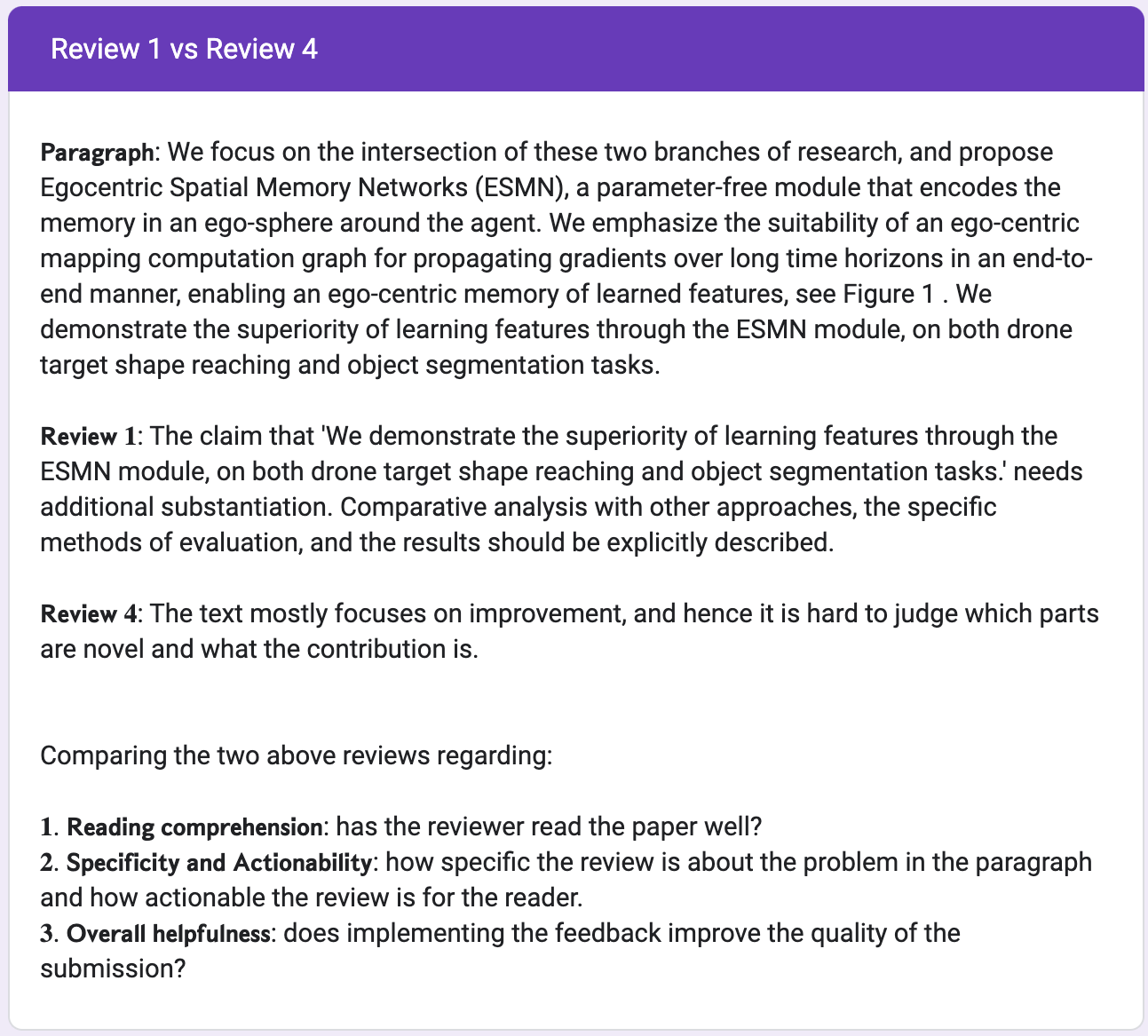}
    \caption{Each comparison involves a paragraph and two reviews. Reviews are randomized at the start of the annotation and no information about the source of the review is provided to prevent bias.}
    \label{fig:paragraph}
\end{figure}
\newpage
 \begin{figure}[H]
    \centering
    \includegraphics[width = .93\linewidth]{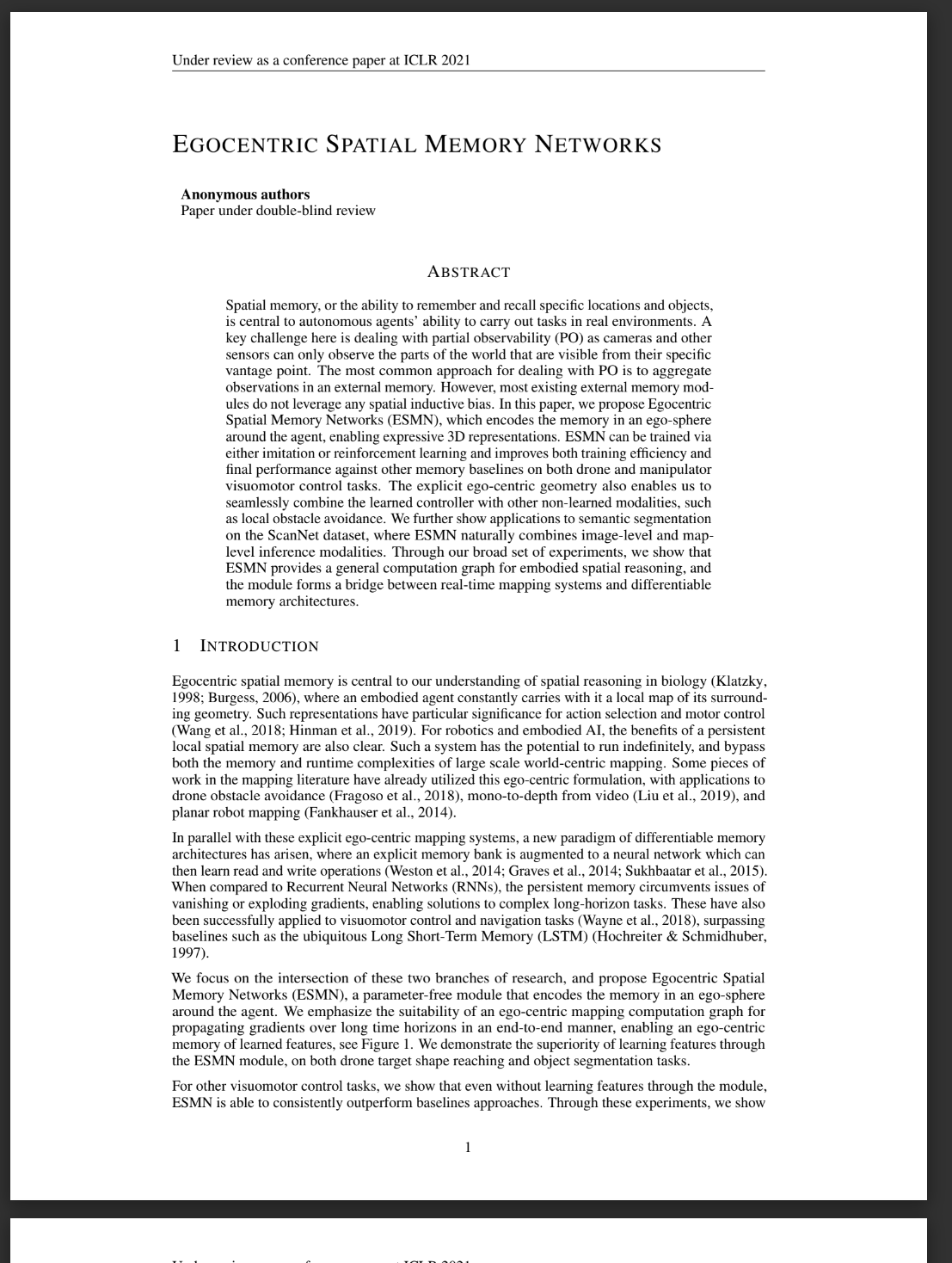}
    \caption{A link is provided to the annotator for convenient access to the version of the paper with the paragraph under review. }
    \label{fig:paper}
\end{figure}
 \begin{figure}[H]
    \centering
    \includegraphics[width = \linewidth]{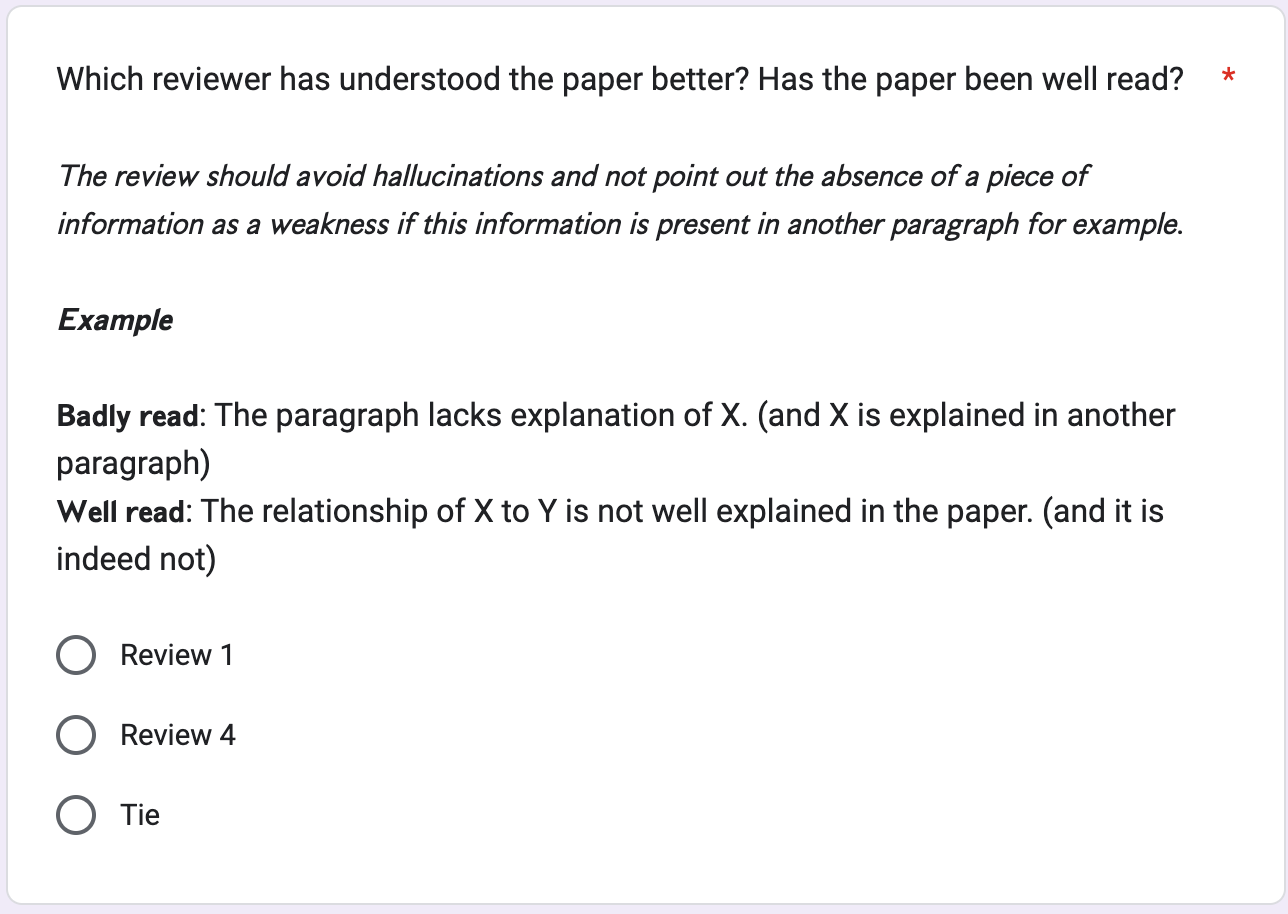}
    \caption{Comparison guidelines for the ``\textit{Reading comprehension}'' criterion. }
    \label{fig:readingcomp}
\end{figure}
 \begin{figure}[H]
    \centering
    \includegraphics[width = \linewidth]{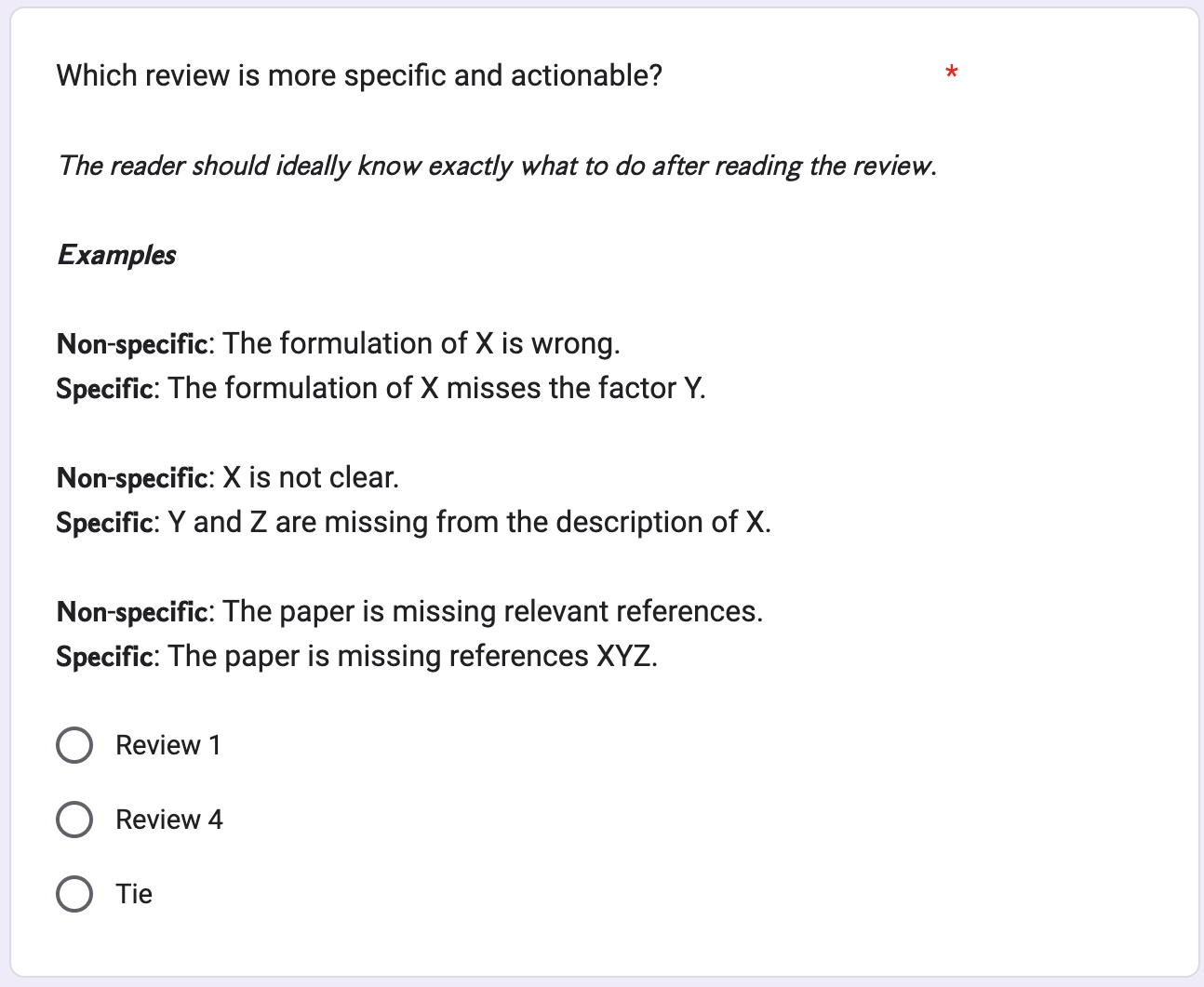}
    \caption{Comparison guidelines for the ``\textit{Specificity and actionability}'' criterion. }
    \label{fig:specificity}
\end{figure}
 \begin{figure}[H]
    \centering
    \includegraphics[width = .93\linewidth]{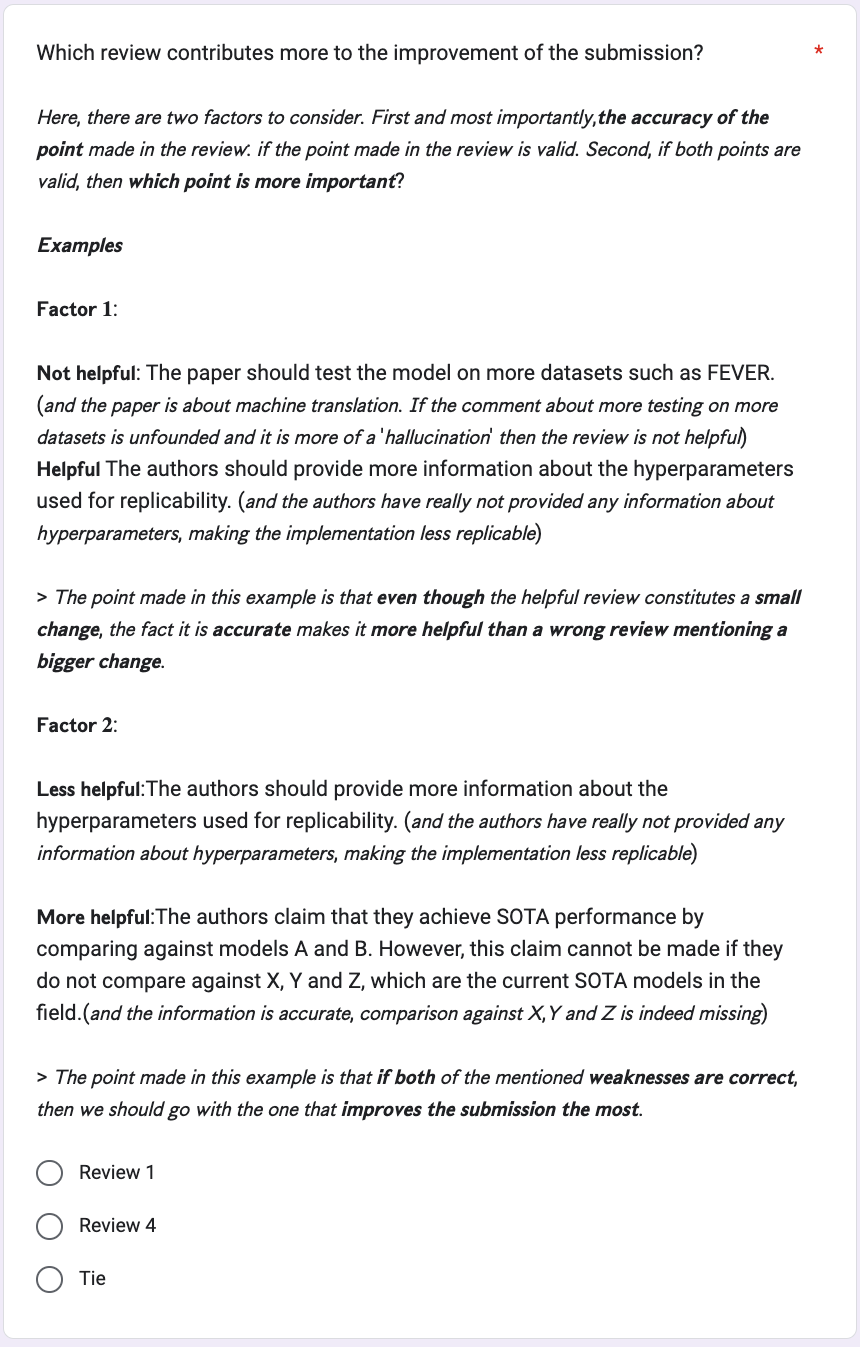}
    \caption{Comparison guidelines for the ``\textit{Overall helpfulness}'' criterion. }
    \label{fig:helpfulness}
\end{figure}

\end{document}